\documentclass[lettersize,journal]{IEEEtran}
\usepackage{amsmath,amsfonts}
\usepackage{algorithmic}
\usepackage{algorithm}
\usepackage{array}
\usepackage[caption=false,font=normalsize,labelfont=sf,textfont=sf]{subfig}
\usepackage{textcomp}
\usepackage{stfloats}
\usepackage{url}
\usepackage{verbatim}
\usepackage{graphicx}
\usepackage{amsmath}
\usepackage{amssymb}
\usepackage{booktabs}
\usepackage{multirow}
\usepackage{array}
\usepackage{graphicx}
\usepackage{float}
\usepackage{color}
\usepackage[section]{placeins}
\usepackage{multirow}
\usepackage{array}
\begin{document}

\title{Motion-Aware Video Frame Interpolation}

\author{\IEEEauthorblockN{Pengfei Han\IEEEauthorrefmark{2}, Fuhua Zhang\IEEEauthorrefmark{2}, Bin Zhao, Xuelong Li,~\IEEEmembership{Fellow,~IEEE}} 
	
	\thanks{

		Pengfei Han is with the School of Cybersecurity and the School of Artificial Intelligence, Optics and Electronics (iOPEN), Northwestern Polytechnical University, Xi'an, Shaanxi 710072, China. 
		
		Fuhua Zhang is with the school of Electrical and Information Engineering, Hunan University, Changsha, 410083, China.
		
		Bin Zhao and Xuelong Li are with the School of Artificial Intelligence, Optics and Electronics (iOPEN), Northwestern Polytechnical University, Xi'an, Shaanxi 710072, China.

\IEEEauthorrefmark{2}{Both authors contributed equally to this work.}

}}

\maketitle

\begin{abstract}
	
Video frame interpolation methodologies endeavor to create novel frames betwixt extant ones, with the intent of augmenting the video's frame frequency.  However, current methods are prone to image blurring and spurious artifacts in challenging scenarios involving occlusions and discontinuous motion. Moreover, they typically rely on optical flow estimation, which adds complexity to modeling and computational costs. To address these issues, we introduce a Motion-Aware Video Frame Interpolation (MA-VFI) network, which directly estimates intermediate optical flow from consecutive frames by introducing a novel hierarchical pyramid module. It not only extracts global semantic relationships and spatial details from input frames with different receptive fields, enabling the model to capture intricate motion patterns, but also effectively reduces the required computational cost and complexity. Subsequently, a cross-scale motion structure is presented to estimate and refine intermediate flow maps by the extracted features. This approach facilitates the interplay between input frame features and flow maps during the frame interpolation process and markedly heightens the precision of the intervening flow delineations. Finally, a discerningly fashioned loss centered around an intermediate flow is meticulously contrived, serving as a deft rudder to skillfully guide the prognostication of said intermediate flow, thereby substantially refining the precision of the intervening flow mappings. Experiments illustrate that MA-VFI surpasses several representative VFI methods across various datasets, and can enhance efficiency while maintaining commendable efficacy.
\end{abstract}


\IEEEpeerreviewmaketitle

\section{Introduction}
As the internet and multimedia technology bloom, video has become a crucial information transmission channel~\cite{kbs1}. 
Currently, a considerable number of videos are recorded at 24/30fps, which is a compromise for hardware devices and communication transmission bandwidth. In practice, high-frame-rate videos provide
better perceptual quality by reducing temporal artifacts~\cite{kbs2, kbs3}. In this case, there is an urgent requirement to get better visual effects by increasing the video frame rate.

Video Frame Interpolation (VFI) undertakes the interpolation of motional data, thereby enhancing the intricacy of alterations traversing from the antecedent frame to the subsequent one.
Capitalizing upon this advantage,
VFI emerges as a versatile tool across an array of video applications, encompassing the augmentation of video frame rates \cite{EA-NET},  the realm of classification \cite{kbs4,kbs5}, and the purview of segmentation~\cite{kbs6,kbs7}.
On the whole, VFI assumes a pivotal role within the realm of computer vision.

\begin{figure}[h]
	\centering
	\includegraphics[width=0.5\textwidth]{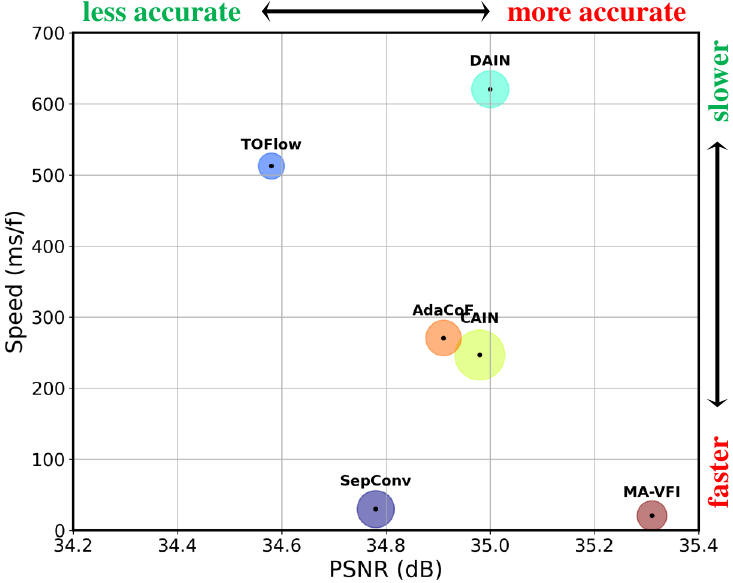} 
	\caption{Running speed \emph{vs}. PSNR values  for different VFI methods, \emph{e.g.}  TOFlow~\cite{xue2019video}, DAIN~\cite{bao2019depth}, AdaCoF~\cite{lee2020adacof}, CAIN~\cite{choi2020channel}, SepConv~\cite{niklaus2017video1}, and proposed MA-VFI on UCF101 dataset. The magnitude of each circle signifies the quantity of model parameters it encompasses.}
	\label{fig:111}
\end{figure}

\begin{figure}[ht]
	\centering
	\includegraphics[width=0.35\textwidth]{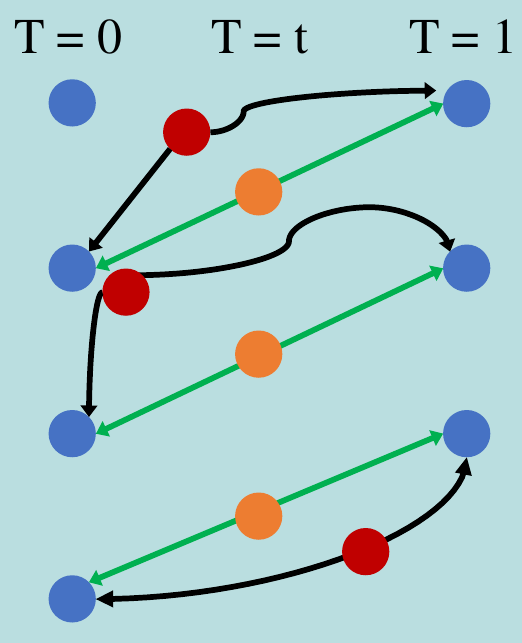} 
	\caption{Visual depiction of pixel motion amidst video frames. The blue circular indicators denote the pixel coordinates corresponding to neighboring video frames at temporal instance T=0 and T=1, while the tangerine markers embody the pixel coordinates of the intermediary frame computed under the presumption of linear motion. In contrast, the crimson markers encapsulate the pixel coordinates of the authentic intermediary frame at time $t$.}
	\label{fig:222}
\end{figure}

Extant VFI techniques have been comprehensively categorized into three primary groups: kernel-based~\cite{niklaus2017video1, reda2018sdc}, phase-based \cite{meyer2015phase, meyer2018phasenet}, and flow-based \cite{xing2021flow, lee2022enhanced} methodologies.
The underlying essence of these methodologies encompasses two fundamental constituents: the estimation of motion and the synthesis of frames. Motion estimation occupies a pivotal role within the domain of VFI, as it substantively dictates the caliber of intermediary frames. The kernel-centric methodology integrates the tasks of motion estimation and frame synthesis as a coherent procedure, yielding the creation of intervening frames. In contrast, phase-based techniques posit that the phase displacement of individual pixel hues can eloquently encapsulate the motion of entities within video frames. Unlike the aforementioned paradigms, flow-based approaches harness optical flow to mathematically decipher motion across video frames. Motion is initially gauged by an optical flow network, subsequently furnishing the basis for the composition of intervening frames through the manipulation of the computed intermediate flow charts.

\subsection{Motivation and Overview}

Recently, VFI approaches made great progress. However, many flow-based VFI methods~\cite{bao2019depth,park2020bmbc} assume linearity in the optical flow field, thereby often engendering one or more of the ensuing constraints: 1)
\emph{\textbf{Temporal Expense of Computation:}} Flow-based strategies predominantly depend upon optical flow and pixel-level warping methodologies. While these algorithms have achieved commendable frame interpolation performance. Finding a compromise between performance and efficiency remains challenging for their implementation in end-to-end applications, as illustrated in Fig.~\ref{fig:111}. 2) \emph{\textbf{Frame Blur and Artifacts:}} Owing to the intricate and multifaceted interplay of forces governing the movement of objects, the underlying assumptions of linear and accelerated motion have certain limitations when it comes to intermediate optical flow estimation. Additionally, in real-world scenarios, factors such as significant displacement of small objects, occlusions, and variations in lighting conditions contribute to issues like motion discontinuity, blurriness, and ghosting artifacts, as can be seen in Fig.~\ref{fig:222}.

Motivated by this, a Motion-Aware Video Frame Interpolation (MA-VFI) network is proposed for VFI task, which can attain a commendable equilibrium between visual quality and the speed of inference. It directly estimates the intermediate flow from consecutive frames, allowing for a more accurate representation of non-linear motion in the real world. Additionally, to effectively capture large-scale motion in video frames and overcome the limitations of receptive fields in convolutional neural networks, MA-VFI utilizes two pyramid modules to estimate multi-scale intermediate flow maps at various receptive fields. Firstly, a pyramid feature module is employed to extract both global semantic relations and spatial detail features from the given frames, enabling the capture of significant motions exhibited by small objects and subtle motions exhibited by larger objects in real-world scenarios. Secondly, a cross-scale motion structure is employed to estimate and refine intermediate flow. It is noteworthy that the features and flow maps between the different layers interact with each other. Specifically, it first employs the higher features to estimate the current intermediate flow maps, then warps the lower features by the predicted flow maps for spatial alignment.
Lastly, the warped features and flow maps are utilized to estimate the next level intermediate flow, which can enhance the interactive ability between features and flow maps in interpolating. Additionally, an intermediate flow-directed loss is specifically designed to guide the estimation of intermediate flow. By employing this approach, it can further assist the model in mitigating issues such as image blurring and spurious artifacts. In summation, the devised MA-VFI methodology attains a remarkable equilibrium between efficiency and efficacy.

\subsection{Contributions}

The principal contributors to MA-VFI implementation could be briefly summarized below:

\begin{itemize}
\item MA-VFI presents a novel hierarchical pyramid feature interpolation model which directly estimates the intermediate optical flow diagram between adjacent frames, alleviating the complex modeling and computational costs of existing methods.

\item  MA-VFI employs cross-scale motion intermediate flow estimation model to enhance the interaction between features and flow maps during the interpolation process, thus improving the representation of nonlinear motion in consecutive frames.

\item The intermediate flow directional loss has been designed to precisely guide the estimation of intermediate flow, thereby assisting the model in alleviating issues such as image blurring and spurious artifacts.

\item	The proposed MA-VFI attains superior results on four publicly accessible datasets than 31 SOTA methods, affirming that it directly attains the utmost equilibrium between performance and efficiency. 
\end{itemize}

\section{Related Work}~\label{Related Work}
VFI is an essential task, which improves video quality by interpolating a single frame or multiple frames. Prevalent methodologies are predominantly clustered into ensuing two main subcategories, namely, kernel- and flow-based. The subsequent subsections expound upon aforementioned approaches.

\subsection{Kernel-based Approaches}
The Kernel-based approaches leverage CNN to combine estimating motion and synthesizing frames in one step for generating intermediate frames. 
Specifically, Niklaus~\emph{et al.} \cite{8099727} have approximated 2D separable convolution kernels on individual synthetic pixel, producing intermediary frames through the convolution of the derived kernels with the input frame.
However, when synthesizing high-resolution video frames, the output kernels require a substantial memory cost. To promote memory efficiency, Niklaus \emph{et al.} \cite{niklaus2017video1} propose to reduce training parameters by separating 2D convolution kernels into 1D. 
DsepConv \cite{cheng2020video} employs deformable separable
convolution network to generate intermediate frames.
CAIN \cite{choi2020channel} adopts multiple channels to get movement information.
Furthermore, AdaCoF~\cite{lee2020adacof} highlights a concern pertaining to the Degrees of Freedom within VFI methodologies. Accordingly, it introduces the concept of generating intermediate frames.
To reduce the parameters of VFI model, Ding \emph{et al.} \cite{ding2021cdfi} propose to compress the AdaCoF model for fast inference. 
CDFI \cite{ding2021cdfi} and LBEC \cite{ding2022video} adopt a lightweight network architecture based on AdaCoF. 
Their results show that the compressed models perform comparable to the AdaCoF. Recently, EDSC~\cite{cheng2021multiple} establishes a linkage to attain the positional information of each pixel. On the other hand, Niklaus and colleagues~\cite{niklaus2022splatting} introduce a methodology for generating intermediary frames without the necessity for subsequent refinement. However, kernel-based techniques inherently lack the capability to directly interpolate numerous intermediary frames between two input frames. Despite the possibility of recursively feeding the intermediary frames  recursively interpolating frames into models to generate multiple intermediary frames, such methodology could potentially result in the accumulation of errors.

\subsection{Flow-based methods}

Flow-based methodologies infer motion by analyzing the optical flow between the given frames. For example, Tu \emph{et al.}  \cite{tu2019survey} furnish a methodical examination of optical flow methodologies rooted in convolutional neural networks.
Tian \emph{et al.}  \cite{tian2020unsupervised} introduce an innovative unsupervised technique for optical flow estimation. Optical flow assumes a significant role within the interpolation procedure. By exerting a direct influence, they significantly impact the quality level of the constructed frames. Consequently, the primary aim of flow-based methodologies is to yield a precise estimation of optical flow through diverse network architectures.
For instance, Liu \emph{et al.} \cite{liu2017video} put forth the Deep Voxel Flow (DVF) model where a 3D network to compute optical flow and synthesize frames with  trilinear sampling. A two-U-Net estimation and refinement method is employed in \cite{jiang2018super}, which can support multi-frame interpolation using bidirectional optical flow estimation and refinement. Considering that the context feature can provide other information besides motion information,  
Niklaus \emph{et al.} \cite{niklaus2018context} propose to warp input frames into intermediate frames with pixel-wise contextual information. 
Xiang \emph{et al.} \cite{xiang2020zooming}  fuse video high-resolution and  interpolation tasks to enhance video frames.
While most existing methods use pre-trained model to improve optical flow accuracy,
TOFlow \cite{xue2019video} uses a spatial pyramid optical flow network \cite{ranjan2017optical} to get optical flow information.
Xue \emph{et al.} \cite{xue2019video} and Bao \emph{et al.} \cite{bao2019depth} use PWC-Net \cite{sun2018pwc} to directly estimate optical flow, whereas MEMC-Net \cite{bao2019memc} employs FlowNet \cite{dosovitskiy2015flownet}. Recently, Niklaus \emph{et al.} \cite{niklaus2020softmax} propose to utilize forward warping to construct intermediate frames with softmax splatting. Xu \emph{et al.} \cite{NEURIPS2019_d045c59a} propose a quadratic video interpolation method that leverages acceleration information in videos to address the challenge of existing video interpolation methods failing to approximate complex real-world motions effectively. The algorithm demonstrates significantly better performance compared to existing linear models. Additionally, Chi \emph{et al.} \cite{Chi2020AllAO} achieve multi-frame video interpolation by introducing a novel flow estimation procedure with a relaxation loss function and a cubic motion model. However, this algorithm struggles to strike a balance between performance and efficiency. Choi \emph{et al.} \cite{9423029} contend that the occurrence of ghosting or tearing artifacts in video interpolation tasks is attributed to the lack of reliable information provided solely by two frames. To address this, they propose an intra-frame interpolation method that obtains tri-directional interpolation information from three input frames. However, the real-time performance of this algorithm is suboptimal. Kalluri \emph{et al.} \cite{FLAVR} leverage three-dimensional spatiotemporal kernels to directly learn the motion attributes of unlabeled videos, proposing an end-to-end multi-frame video interpolation algorithm. While these methods have achieved commendable frame interpolation performance, they grapple with the challenge of harmonizing performance and efficiency. Additionally, real-world scenarios encompass factors such as significant displacement of small objects, occlusions, and variations in lighting conditions, which result in issues like motion discontinuity, blurring, and ghosting in the generated intermediate frames.

\section{Motion-Aware Video Frame Interpolation} \label{network}

\begin{figure*}[t]
	\centering
	\includegraphics[width=1.0\textwidth]{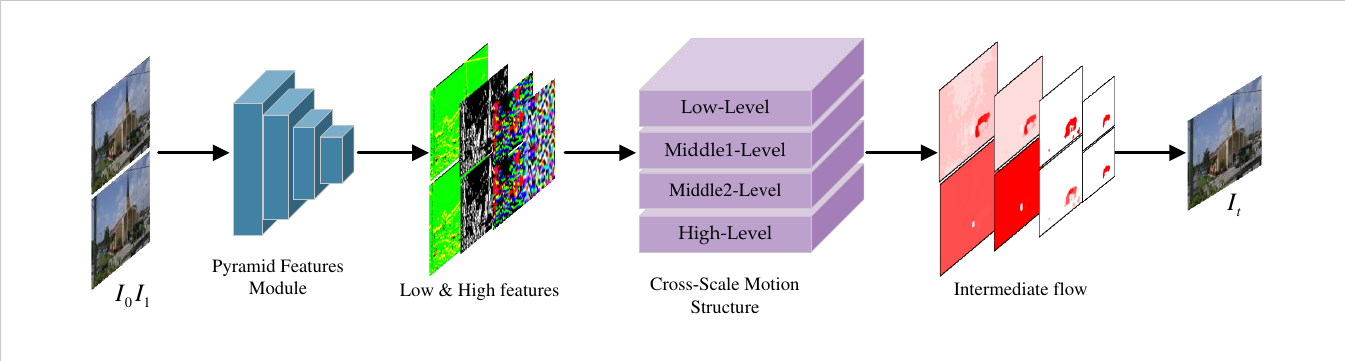} 
	\caption{An Outline of the Motion-Aware Video Frame Interpolation Network, referred to as MA-VFI. This architecture comprises two integral components: the Pyramid Features Module and the Cross-Scale Motion Structure. In the initial stage, low-level and high-level features need to be extracted from the input frames, which subsequently facilitate the computation of intermediate flow maps essential for interpolation.
	}
	\label{fig1}
\end{figure*}

Given a sequential pair of frames \({I_0}\) and \({I_1}\) originating from a video with a low frame rate, the envisaged approach endeavors to generate an intervening frame \(I_t\).
In Fig. \ref{fig1}, the proposed Motion-Aware network (MA-VFI) is illustrated. 
It is structured into two integral components, namely, the hierarchical pyramid feature extraction module, and the cross-scale motion intermediate flow estimation module. Firstly, the pyramid features module is employed to extract global semantic relations and spatial detail features from  \({I_0}\) and \({I_1}\). Then, the extracted features are used to estimate and refine intermediate flow in the cross-scale motion structure. The following is the detailed presentation of each part.

\subsection{Hierarchical Pyramid Feature Extraction}
The existing VFI methods first estimate the intermediate flow maps to generate intermediated frames, and then refine them by input frames features. However, the cascaded architecture ignores their mutual promotion during interpolation. This results in the obfuscation of intricate particulars within the produced intermediary frame. To alleviate aboved issue, MA-VFI forms a close connection between them and promote the mutual assistance between input frames features and flow maps during interpolating frame. 
Specifically, a pyramid feature module is used to extract global semantic relations and spatial detail features from given frames, and then employ those features to estimate intermediate flow maps.

The pyramid feature module includes four layers, which are employed to extract comprehensive semantic and spatial details. Each layer is composed of two convolution blocks with step sizes 2 and 1. The different receptive fields are used to capture features in the pyramid module.
The filter sizes are set as 7\(\times\)7 and 3\(\times\)3 in the first layer, the rest layers are all set 3\(\times\)3.
To gradually extract space size, the features channels of each layer are successively increased to 64, 96, 144, 192.
global semantic relations and spatial detail features are extracted from different receptive fields in the pyramid feature module as:         
\begin{equation}\begin{split}
	{F_0^0},{F_0^1},{F_0^2},{F_0^3} =  { \rm{PFM}}  ({I_{0}}),
\end{split}\end{equation}
\begin{equation}\begin{split}
	{F_1^0},{F_1^1},{F_1^2},{F_1^3} =  { \rm{PFM}}  ({I_{1}}),
\end{split}\end{equation}
where \({F_0^i}\) and \({F_1^i}\) represent the contextual features, PFM denotes the pyramid feature model, and $i\in (0,1,2,3)$. 

\subsection{Cross-scale Motion Intermediate Flow Estimation}

Prevalent flow-based methodologies initially compute intermediary flow maps employing either a linear or quadratic motion model, subsequently generating intervening frames through the deformation of provided frames.
However, calculating intermediate flow in a fixed motion model is unable to closely simulate the complex motion of the real-world. The fixed motion model leads to blurring and artifacts in the synthesized intermediate frames.

\begin{figure}
\centering
\includegraphics[width=0.5\textwidth]{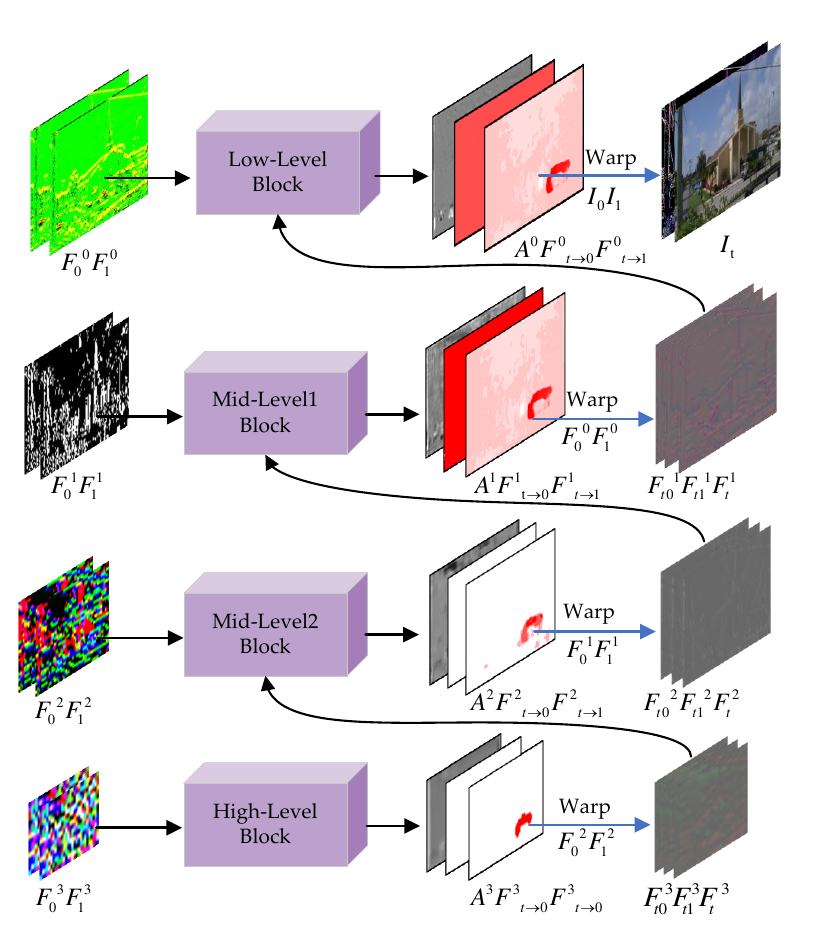}  
\caption{Overview of the cross-scale motion structure. It is a top-down structure with lateral connections. The structure computes intermediate flow maps at four scales and refines them with residual flows at each level.}
\label{fig2}
\end{figure}

 Motivated by this,
the proposed MA-VFI directly forecasts intermediary flow \({F_{t\rightarrow0}}\) and \({F_{t\rightarrow1}}\)  for  the undertaking of VFI.   
It is more adept at delineating the intricate non-linear motion within the preceding and subsequent frames compared to the rigid motion model.
Specifically, a cross-scale motion structure is designed to estimate intermediate flow.
As depicted in Fig. \ref{fig2}, 
it is a top-down structure with lateral connections. 
In this structure, the intermediate flow maps are computed at four scales and refined with
residual flow maps at each level. It includes four Intermediate Flow Blocks (IFBlocks), \emph{i.e.}, the low-level block, the mid-level1 block, the mid-level2 block, and the high-level block. The architecture ascertains intermediary flow and guide maps through a progression from rough to refined. The inputs for each Intermediate Flow Block comprises the extracted features, yielding the outputs of intermediary flow and guide maps.
In particular, the upper-level features are initially leveraged to approximate intermediary flow maps in the high-level block. Subsequently, the lower-level features undergo deformation by the prognosticated flow maps to achieve alignment. Eventually, the warped features and prevailing flow maps are furnished to the subsequent level for the computation of intermediary flow. This augments the symbiotic interplay between features and intermediary flow maps during the interpolation process, resulting in high-quality intermediary frames imbued with discernible motion contours and intricate particulars.

Firstly, the input of High-Level Block (HLBlock) are the highest features \({F_0^3}\) and \({F_1^3}\), the output is \({F^3_{t\rightarrow0}}\), \({F^3_{t\rightarrow1}}\), and  \({A^3}\) as:

\begin{equation}\begin{split}
	{F^3_{t\rightarrow0}},{F^3_{t\rightarrow1}}, {A^3}  =  { \rm{HLBlock}}  ({F^{3}_{0}},{F^{3}_{1}}),
\end{split}\end{equation}
where \({F^{3}_{t\rightarrow0}}\), \({F^{3}_{t\rightarrow1}}\), and  \(A^{3}\) are the current estimated intermediate flow and guide maps. To enhance the interactive ability between frames features and intermediate flow maps in interpolating.
The lower features \(F_0^2\), \(F_1^2\) are warped by the predicted flow maps \(F^3_{t\rightarrow0}\), \(F^3_{t\rightarrow1}\) for spatial alignment.
The warped features \({F^3_{t0}}\), \({F^3_{t1}}\), and \({F^3_t}\) can be obtained by warping input frames as:
\begin{equation}
{{F^3_{t0}}}=  {\rm warp} {({F^2_0},{ F^3_{t\rightarrow0}})}, 	{{F^3_{t1}}}=  {\rm warp} {({F^2_1},{ F^3_{t\rightarrow1}})}, \label{frame features}
\end{equation} 
\begin{equation}
{{F^3_{t}}}=  {{F^3_{t0}}}\odot{A^{3}} + {{F^3_{t1}}}\odot{(1-A^{3})}, \label{Intermediate features}
\end{equation}
where $\odot$ signifies an element-wise multiplier, \(A^3\) is the guide map, which combines the information from two images.

After getting the initial results from HLBlock,
we gradually refine intermediate flow maps through Mid-Level2 Block (MLBlock2), Mid-Level1 Block (MLBlock1), and Low-Level Block (LLBlock). Specifically, the   \({F^2_{0}}\), \({F^2_{1}}\), \({F^3_{t0}}\), \({F^3_{t1}}\), and \({F^3_t}\) are connected together and entered into MLBlock2. The outputs of MLBlock2 are the refined results \({F^2_{t\rightarrow0}}\), \({F^2_{t\rightarrow1}}\), and  \({A^2}\) as :
\begin{equation}\begin{split}
	{F^2_{t\rightarrow0}},{F^2_{t\rightarrow1}}, {A^2}  & =  { \rm{MLBlock2}}  ({F^{2}_{0}},{F^{2}_{1}}, {F^{3}_{t0}},{F^{3}_{t1}},{F^3_{t}})  + \\ & {\rm{Up}({F^3_{t\rightarrow0}},{F^3_{t\rightarrow1}})} ,
\end{split}\end{equation}
where "Up" denotes the process of upsampling, which is employed to magnify the acquired intermediary flow for the purpose of concatenating residuals and refining the flow maps.
As with the MLBlock2, the intermediary flow \({F^1_{t\rightarrow0}}\), \({F^1_{t\rightarrow1}}\) and guide map \(A^1\) are gotten from MLBlock1. The warped features  \({F^1_{t0}}\), \({F^1_{t1}}\), \({F^1_t}\) are also obtained.
Note that the output of LLBlock encompasses
the ultimate intermediary flow  \({F_{t\rightarrow0}}\), \({F_{t\rightarrow1}}\), guide map \(A\), and residual map \(R\). They follow formulation: 
\begin{equation}\begin{split}
	{F_{t\rightarrow0}},{F_{t\rightarrow1}},{A},{R} & =  { \rm{LLBlock}}  ({F^{0}_{0}},{F^{0}_{1}}, {F^{1}_{t0}},{F^{1}_{t1}},{F^1_{t}})  + \\ & {\rm{Up}({F^1_{t\rightarrow0}},{F^1_{t\rightarrow1}})}.
\end{split}\end{equation}

Lastly, the intermediate frame \({I_t}\) is obtained as:
\begin{equation}
{{I_{t\rightarrow0}}}=  {\rm warp} {({I_0},{ F_{t\rightarrow0}})}, 	{{I_{t\rightarrow1}}}=  {\rm warp} {({I_1},{ F_{t\rightarrow1}})}, 
\end{equation}
\begin{equation}
{{I_{t}}}=  {{I}{_{t\rightarrow0}}}\odot{A} + {{I_{t\rightarrow1}}}\odot{(1-A)} +{R}. 
\end{equation}

\begin{figure}
\centering
\includegraphics[width=0.5\textwidth]{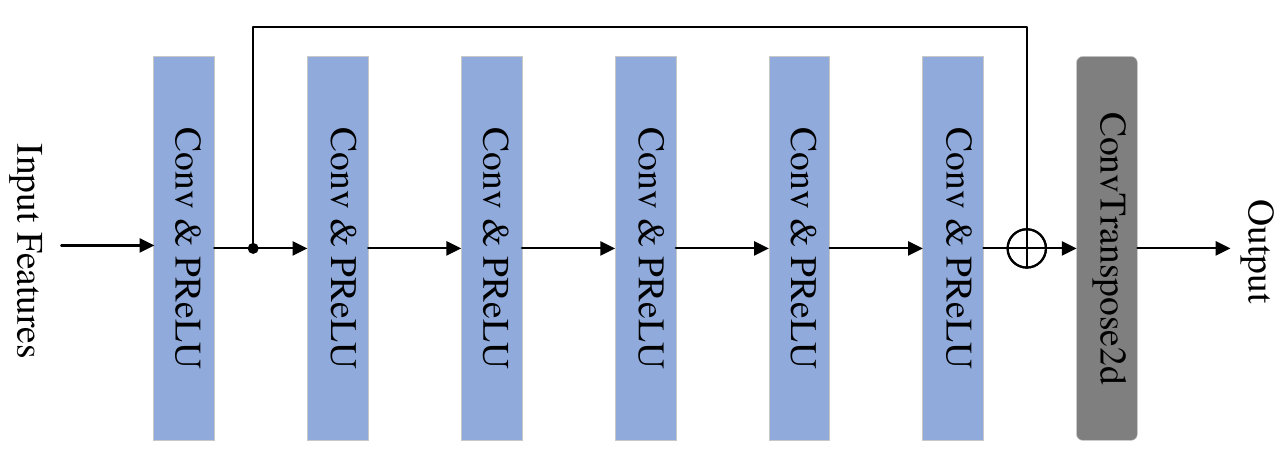} 
\caption{Sketch of intermediate flow block architecture. Note that $\oplus$ represents an element-wise multiplier.}
\label{fig3}
\end{figure}

To make MA-VFI lightweight and arrive real-time video frame interpolation tasks, 
a simple but powerful 3 \(\times\) 3 Conv layer with PReLU activation is used to form cross-scale motion structure. 
Four intermediate flow blocks have the same network, which is demonstrated in Fig. \ref{fig3}. 
Each block consists six Conv layer with strides 1 and one 4 \(\times\) 4 deconvolution.
Moreover, it adopts residual connection to promote information propagation.
Pyramid feature modules and cross-scale motion structures make up the MA-VFI. This encompasses a comprehensive end-to-end network meticulously crafted with the specific intention of interpolating video frames.

\subsection{Optimization Strategy}

The proximity  among synthesized frames and the real frame is assessed through the metric of reconstruction loss \(\mathcal{L}_{rec}\).
It makes the synthesized frames sharper and keeps more detailed information of the synthesized frames. The \(\mathcal{L}_{rec}\) loss is determined by the \(L_1\) norm. 
It is formulated as follows:
\begin{equation}{\mathcal{L}_{rec}} = {\parallel{{{({I^{GT}_t}}{-}{I_t}})}\parallel}_1, \end{equation}
where \(I^{GT}_{t}\) signifies the actual intermediary frame, and the \(I_{t}\) represents the predicted intermediary frame. Furthermore, to make adjacent pixels have similar flow values, the smoothness loss \cite{liu2017video} is employed as:

\begin{equation}{\mathcal{L}_{smooth}} = {\parallel{\bigtriangledown}{F^i_{t\rightarrow0}}\parallel}_1 + {\parallel{\bigtriangledown}{F^i_{t\rightarrow1}}\parallel}_1 , i = (0,1,2,3),\end{equation}
where \(F^i_{t\rightarrow0}\) and \(F^i_{t\rightarrow1}\) denote the intermediate flow from each IFBlock.  

Although training MA-VFI using \(\mathcal{L}_{rec}\) and \(\mathcal{L}_{smooth}\) can generate intermediate frames. However, simple loss functions tend to fall into the problem of local minima. In addition, taking into account the absence of supervision for intermediate flow in the input to MA-VFI, 
a loss designated as \(\mathcal{L}_{flow}\), governed by the intermediate flow, is formulated.

Specifically, a pre-trained LiteFlownet~\cite{hui2018liteflownet} is used to get the multi-scale directed intermediate optical flow \(F^d_{t\rightarrow0}\) and \(F^d_{t\rightarrow1}\), which play a pivotal role in advancing the prediction of intermediate motion. Inspired by \cite{huang2022real},
the intermediate flow-directed loss \(\mathcal{L}_{flow}\) can be written as :
\begin{equation}
{\mathcal{L}_{flow}} = {\parallel{{{({F^{d^{}}_{t\rightarrow0}}}{-}{F^{i}_{t\rightarrow0}}})}\parallel}_1+
{\parallel{{{({F^{d^{}}_{t\rightarrow1}}}{-}{F^{i}_{t\rightarrow1}}})}\parallel}_1,
\end{equation}
where \(F^i_{t\rightarrow0}\) and \(F^i_{t\rightarrow1}\) represent the intermediate flow maps on different levels. The expression for the ultimate comprehensive loss term is:
\begin{equation}
\mathcal{L}= {\alpha}{\mathcal{\mathcal{L}}_{rec}} + {\beta}{\mathcal{L}_{flow}}+{\gamma}{\mathcal{L}_{smooth}}.
\end{equation}
where the $\alpha$, $\beta$, and $\gamma$ are the weighting factors. $\mathcal{L}_{rec}$ denotes the reconstruction loss,  $\mathcal{L}_{flow}$ represents the intermediate flow-based loss, and the $\mathcal{L}_{smooth}$ is the smooth loss.

\section{Experiment} \label{Experiment}
Within this section, MA-VFI undergoes examination and scrutiny across both single- and multi-VFI endeavors. Initially, the datasets and particulars of training are delineated. Subsequently, a juxtaposition between MA-VFI and notable methodologies is conducted over diverse benchmarks. Lastly, ablation analyses are conducted to validate the enhancements attributed to the introduced MA-VFI technique.
\begin{figure*}[ht]
\centering
\includegraphics[width=1.0\textwidth]{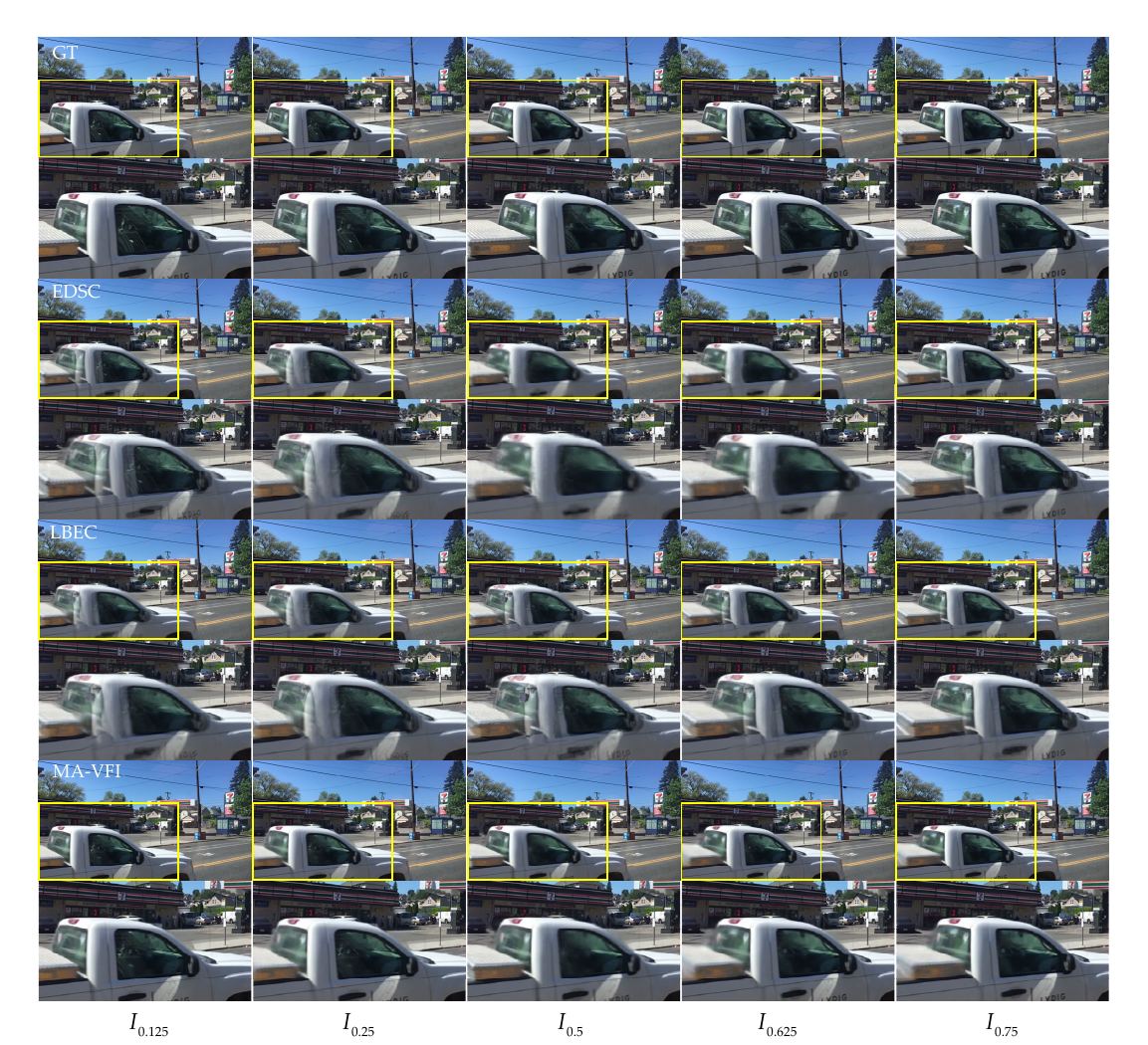} 
\caption{Visual juxtaposition of the synthesized intermediate frames on the Adobe240fps dataset with different methods. The first and second row display the ground truth and the zoomed details of the yellow rectangular area, respectively. The other rows display the five intermediate frames generated by the different methods. 
	In each column from left to right are the intermediate frames generated at different timesteps.}
\label{fig6}
\end{figure*}
\subsection{Experimental Settings} 
\subsubsection{Datasets}
In this paper, four well-known and challeng-
ing datasets have been employed to evaluate the efficacy of the suggested methodology as follows.

\begin{figure*}[h]
\centering
\includegraphics[width=0.9\textwidth]{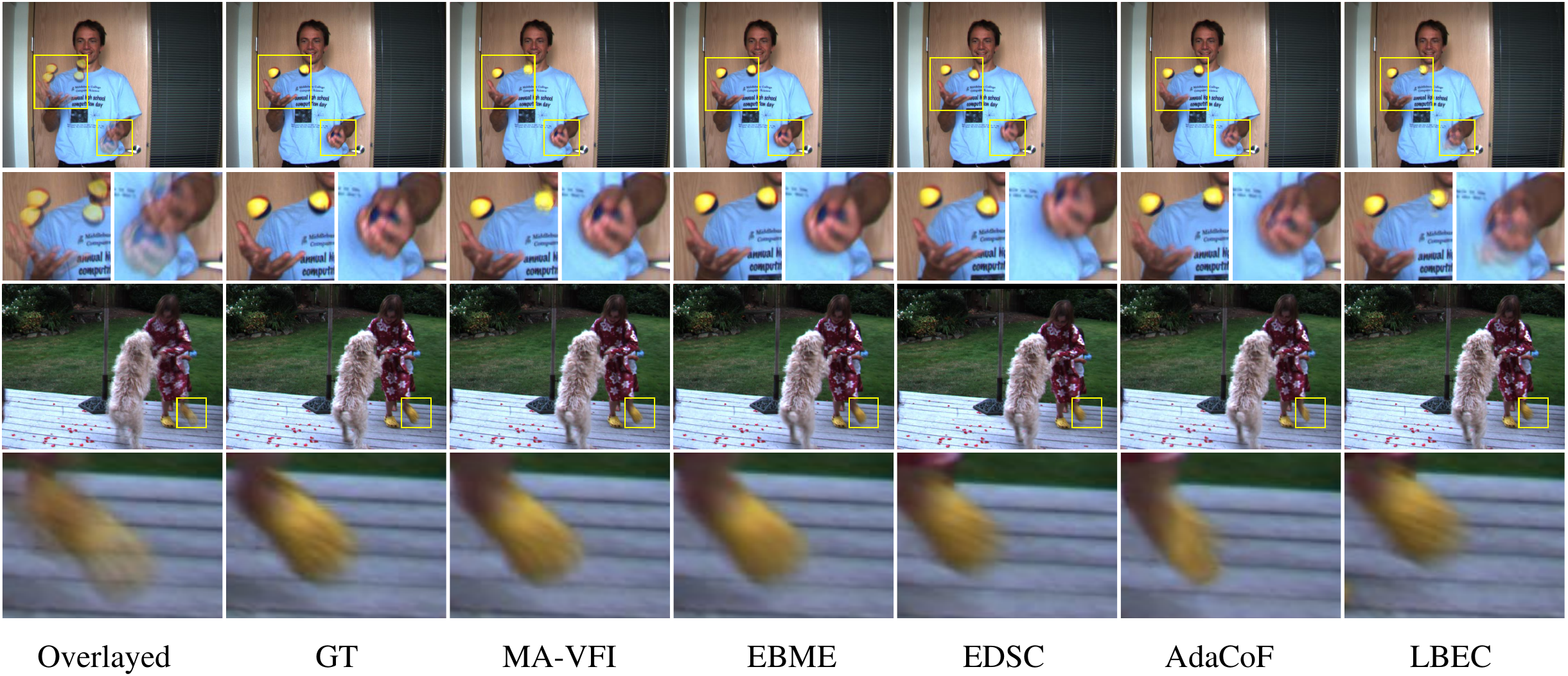} %
\caption{Visual contrast of the generated intermediate frames using diverse methodologies on the Middlebury dataset. The initial and second rows showcase the reference truth of intermediate frames and magnified specifics of the yellow rectangular region, respectively. The subsequent rows exhibit the five intermediate frames created by distinct techniques. Each column, progressing from left to right, corresponds to the intermediate frames produced at varying time intervals.}
\label{fig5}
\end{figure*}

\textbf{Vimeo-90k Dataset}: Vimeo-90k \cite{xue2019video}
serves as the training dataset for MA-VFI. Comprising a total of 91,701 triplets, the training subset encompasses 51,313 triples, leaving the remaining 3,782 triples for validation and testing purposes. To elaborate,  each image frame possesses dimensions of 448 × 256, and every sequence of three successive frames constitutes a triplet.

\textbf{UCF101 Dataset}: 
The UCF101 dataset \cite{soomro2012dataset} encompasses a diverse array of human behaviors captured across various settings. This dataset comprises 379 triplets, each exhibiting a resolution of 256
×256 pixels. Specifically, it was employed for evaluating single-frame interpolation.

\textbf{Middlebury Dataset}: 
The Middlebury dataset \cite{ha2004motion} is harnessed for single-frame interpolation evaluations. Renowned for its role in appraising optical flow and VFI endeavors, this dataset comprises 12 pairs of images, each boasting a resolution of approximately 640
×480 pixels.

\textbf{Adobe-240fps Dataset}: The Adobe-240fps dataset \cite{su2017deep} encompasses 112 video clips characterized by a resolution of 720 × 1280. This compilation comprises 254 sequences, each comprising 9 frames. Notably, the initial and concluding frames within these sequences are utilized to assess the reliability of the multi-frame interpolation task.

\subsubsection{Training details}

Throughout the training phase on the Vimeo-90k dataset, our MA-VFI model is optimized using the AdamW optimizer \cite{kingma2014adam}. It is with weight decay \({10^{-4}}\) for 300 epochs, while maintaining a batch size of 6. The learning rate undergoes a gradual reduction from \({3{\times}10^{-4}}\) to \({3{\times}10^{-5}}\), following a cosine annealing strategy throughout the entirety of the training protocol.
The implementation employs PyTorch \cite{paszke2019pytorch} on GeForce RTX 3090 GPUs for the training of MA-VFI.

\begin{table*}[h]
			\renewcommand\arraystretch{0.9}
			\caption{ Effectiveness of single-frame video interpolation
				across the UCF101, Vimeo-90k and Middlebury datasets.}
			\centering
			\setlength\tabcolsep{8.0pt}
			\begin{tabular}{cccccccccc}
				\noalign{\hrule height 1.1pt}
				\hline
				\multirow{2}{*}{ \textbf{Methods}} &\multirow{2}{*}{ \textbf{Publications}} & \multicolumn{2}{c}{\textbf{UCF101}}        & \multicolumn{2}{c}{\textbf{Vimeo-90k}}     & {\textbf{Middlebury}}  & {\textbf{Parameters}}  & {\textbf{Runtime}}  \\
				\cmidrule(r){3-4}  	\cmidrule(r){5-6} \cmidrule(r){7-7}\cmidrule(r){8-8}\cmidrule(r){9-9}
			&	&  \multicolumn{1}{c}{\textbf{PSNR}}   & {\textbf{SSIM}}  & \multicolumn{1}{c}{\textbf{PSNR}}   & {\textbf{SSIM}}   & {\textbf{IE}}  &{\textbf{(Million)}}  & {\textbf{(s)}}     \\ 
				\hline	
\multicolumn{8}{c}{\textbf{Kernel-based}} \\ \hline
	MIND \cite{MIND}           & ECCV2016 & \multicolumn{1}{c}{33.93} & 0.966 & \multicolumn{1}{c} {33.50} & 0.942 &3.35&------  &0.064     \\
SepConv \cite{niklaus2017video1}     &  ICCV2017     & \multicolumn{1}{c}{34.78} & 0.967 & \multicolumn{1}{c}{33.79} & 0.970 & 2.27 & 21.68 & 0.030     \\ 
 CAIN \cite{choi2020channel}      & AAAI2020    & \multicolumn{1}{c}{34.98} & {{\textbf{\textcolor{blue}{0.969}}}} & \multicolumn{1}{c}{34.65} & 0.973 & 2.28 & 42.78 & 0.247       \\ 		
DsepConv  \cite{cheng2020video}      &  AAAI2020   & \multicolumn{1}{c}{35.08} & {{\textbf{\textcolor{blue}{0.969}}}} & \multicolumn{1}{c}{34.73} & 0.974 & 2.06     & 21.8 & 0.138     \\ 
EDSC \cite{cheng2021multiple}   & TPAMI2021 & \multicolumn{1}{c}{35.13} & \textbf{0.968} & \multicolumn{1}{c}{34.84} & 0.975 & 2.02 & 8.9 & {{0.046}}    \\ 
 CDFI \cite{ding2021cdfi}       &  CVPR2021    & \multicolumn{1}{c}{35.21} & {{\textbf{\textcolor{blue}{0.969}}}} & \multicolumn{1}{c} {35.17} & 0.977 & ------& 5.0 &0.064      \\ 
\hline
 \multicolumn{8}{c}{\textbf{Flow-based}}\\ \hline
DVF \cite{liu2017video}         & ICCV2017       & \multicolumn{1}{c}{34.92} & \textbf{0.968} & \multicolumn{1}{c}{34.56} & 0.973 & 4.04 & ------ &  0.043    \\ 
 Superslomo \cite{jiang2018super}    & CVPR2018  & \multicolumn{1}{c}{35.15} & \textbf{0.968} & \multicolumn{1}{c}{34.64} & 0.974 & 2.51 & 39.6 & \textbf{0.023}     \\ 
 CtxSyn \cite{ctxsyn}  &CVPR2018  & \multicolumn{1}{c}{34.01} & {0.941} & \multicolumn{1}{c}{{33.76}} & 0.955  &2.17 & ------  & ------   \\ 
TOFlow  \cite{xue2019video}  & IJCV2019   & \multicolumn{1}{c}{34.58} & 0.967 & \multicolumn{1}{c}{33.73} & 0.968 & 2.15 & 1.07 & 0.513      \\ 
DAIN   \cite{bao2019depth}      &  CVPR2019       & \multicolumn{1}{c}{35.00} & \textbf{0.968} & \multicolumn{1}{c}{34.71} & 0.976 & 2.04 & 24.02& 0.621     \\ 
 BMBC \cite{park2020bmbc}       &   ECCV2020   & \multicolumn{1}{c}{35.15} & {{\textbf{\textcolor{blue}{0.969}}}} & \multicolumn{1}{c} {35.01} & 0.975 & 2.04 & 11.0 &1.603      \\ 
 Softsplat-$Lap$ \cite{niklaus2020softmax}       &   CVPR2020   & \multicolumn{1}{c}{35.10} & 0.948 & \multicolumn{1}{c} {35.48} & 0.964 & ------ & 7.7 & 0.135      \\ 
Softsplat-$F$ \cite{niklaus2020softmax}       &   CVPR2020   & \multicolumn{1}{c}{{{\textbf{\textcolor{red}{35.39}}}}} & 0.952 & {{\textbf{\textcolor{blue}{36.10}}}} & 0.970 & ------   & 12.2 & 0.195    \\ 
XVFI \cite{xvfi} &ICCV2021 &35.18&0.952&35.07&0.968&------ &5.5&0.098 \\
ACTA  \cite{ACTA} &ICIP2022 &35.23&{{\textbf{\textcolor{blue}{0.969}}}}&35.90&{{\textbf{\textcolor{blue}{0.979}}}}&2.01 &3.4&0.056 \\
 RIFE \cite{huang2022real}   & ECCV2022   & 35.28 & {{\textbf{\textcolor{blue}{0.969}}}} & 35.61 & \textbf{0.978}  &{{\textbf{\textcolor{blue}{1.96}}}} & 9.8 & {{\textbf{\textcolor{red}{0.016}}}}    \\ 
 M2M-PWC \cite{TOmany}   & CVPR2022  & \multicolumn{1}{c}{{35.17}} & {{\textbf{\textcolor{red}{0.970}}}} & \multicolumn{1}{c}{{35.40}} & \textbf{0.978}  &------  &7.6  &0.12   \\ 
M2M-DIS~\cite{TOmany}  & CVPR2022 & \multicolumn{1}{c}{{35.13}} & \textbf{0.968} & \multicolumn{1}{c}{{35.06}} & {0.976}  &------  &------  &------   \\ 
IFRNET-S~\cite{ifrnet} &  CVPR2022  & \multicolumn{1}{c}{35.28} & {{\textbf{\textcolor{blue}{0.969}}}} & 35.59 &\textbf{0.978}  &2.03  &2.8  &------   \\ 
 MFNet~\cite{mfnet}  & TMM2023 & 35.16&0.962 &  35.64 &0.976 & 1.97&------  &0.068 \\
 EBME \cite{10030216}   & WACV2023  & \multicolumn{1}{c}{35.30} & {{\textbf{\textcolor{blue}{0.969}}}} & \multicolumn{1}{c}{{35.58}} & {{\textbf{\textcolor{red}{0.980}}}}  &------ & 3.9 & \textbf{0.027}     \\ 
ProBoost-Net~\cite{ProBoost-Net}  & TMM2023 & {{\textbf{\textcolor{blue}{35.33}}}} &\textbf{0.968} &  {{\textbf{\textcolor{red}{36.18}}}} &{{\textbf{\textcolor{blue}{0.979}}}} & ------ & 12.4 &0.319 \\
DIS-M2M~\cite{M2M}  & TPAMI2024  & \multicolumn{1}{c}{35.13} & \textbf{0.968} & \multicolumn{1}{c}{35.06} & 0.976 & ------ & ------ & 0.028     \\ 
DIS-M2M++~\cite{M2M}  & TPAMI2024  & \textbf{35.31} & {{\textbf{\textcolor{blue}{0.969}}}} & \multicolumn{1}{c}{35.78} & {{\textbf{\textcolor{red}{0.980}}}} & ------ & ------ & 0.278     \\ 
EA-Net \cite{EA-NET}    & TNNLS2024  & \multicolumn{1}{c}{{34.97}} & {0.967} & \multicolumn{1}{c}{{34.39}} & {0.975}  &2.17 & ------ &------     \\
\hline
 \multicolumn{8}{c}{\textbf{Kernel and Flow Combination}} \\ \hline
AdaCoF~\cite{lee2020adacof}        & CVPR2020   & \multicolumn{1}{c}{34.91} & \textbf{0.968} & \multicolumn{1}{c}{34.27} & 0.971 & 2.24 & 21.84& 0.271     \\ 
 MEMC-Net \cite{bao2019memc}      &  TPAMI2021  & \multicolumn{1}{c}{35.01} & \textbf{0.968} & \multicolumn{1}{c}{34.29} & 0.970 & 2.12 & 70.3 & 0.235   \\ 
FGME \cite{FGME} &  TBC2021 & \multicolumn{1}{c}{35.12} & {0.950} & \multicolumn{1}{c}{{34.91}} & 0.964  &2.17 & 11.95 &0.095     \\ 
ReMEI-Net~\cite{hucs}    & TCSVT2022 & \multicolumn{1}{c}{{35.07}} & \textbf{0.968} & \multicolumn{1}{c}{{34.58}} & {0.972}  &2.13 &------  &------   \\				
\hline
 \multicolumn{8}{c}{\textbf{Others}} \\ \hline
\hline
PDWN~\cite{pdwn}    &IOJSP2021  & \multicolumn{1}{c}{{35.00}} & {0.950} & \multicolumn{1}{c}{{35.44}} & {0.966}  &\textbf{1.98}&------  &0.086   \\ 
GDConvNet~\cite{gdcnet}    & TMM2022  & \multicolumn{1}{c}{{35.16}} & \textbf{0.968} & \multicolumn{1}{c}{{34.99}} & {0.975}  &2.03 &5.6  &------   \\ 
LBEC~\cite{ding2022video}  & ICASSP2022  & \multicolumn{1}{c}{35.21} & {{\textbf{\textcolor{blue}{0.969}}}} & \multicolumn{1}{c}{34.58} & 0.973 & ------ & 7.05 &0.030     \\ 
RBFPNet~\cite{RBFPNet}  & IETIP2023  & \multicolumn{1}{c}{32.48} & 0.928 & \multicolumn{1}{c}{34.66} & 0.960 & ------ & 33.63 & ------     \\ 			
\hline		
MA-VFI (Ours)      &     & \textbf{35.31} & {{\textbf{\textcolor{red}{0.970}}}} & 
				\textbf{35.96} & {{\textbf{\textcolor{red}{0.980}}}} & {{\textbf{\textcolor{red}{1.91}}}} & 17.87 & {{\textbf{\textcolor{blue}{0.021}}}}     \\ \hline
				\noalign{\hrule height 1.1pt}
			\end{tabular}\label{table.1}
		\end{table*}

\subsubsection{Assessment criteria}

To gauge the efficacy of our methodology, three comprehensive metrics have been chosen for this experiment, namely, Peak Signal to Noise Ratio (PSNR), Structural Similarity Index (SSIM), and Interpolation Error (IE),
which
have a positive correlation with performance.
In general, a smaller IE value implies better intermediate frames performance,
while higher values of PSNR and SSIM suggest superior visual
quality of the intermediate frames.

\subsubsection{Contrasting methods}

Within this section, to appraise the effectiveness of the presented MA-VFI method against other approaches, we conducted a comparative analysis of SOTA VFI methods, including MIND \cite{MIND}, SepConv \cite{niklaus2017video1}, CAIN \cite{choi2020channel}, DsepConv \cite{cheng2020video}, EDSC \cite{cheng2021multiple}, CDFI \cite{ding2021cdfi}, DVF \cite{liu2017video}, Superslomo \cite{jiang2018super}, CtxSyn \cite{ctxsyn}, TOFlow \cite{xue2019video}, DAIN \cite{bao2019depth}, BMBC \cite{park2020bmbc},  Softsplat-$Lap$ \cite{niklaus2020softmax}, Softsplat-$F$ \cite{niklaus2020softmax}, XVFI \cite{xvfi}, ACTA  \cite{ACTA}, RIFE \cite{huang2022real}, M2M-PWC \cite{TOmany}, M2M-DIS~\cite{TOmany}, IFRNET-S~\cite{ifrnet}, MFNet~\cite{mfnet}, EBME \cite{10030216}, ProBoost-Net~\cite{ProBoost-Net}, DIS-M2M~\cite{M2M}, DIS-M2M++~\cite{M2M}, EA-Net \cite{EA-NET}, AdaCoF~\cite{lee2020adacof}, MEMC-Net \cite{bao2019memc}, FGME \cite{FGME}, ReMEI-Net~\cite{hucs}, 
PDWN~\cite{pdwn}, GDConvNet~\cite{gdcnet}, LBEC~\cite{ding2022video} and RBFPNet~\cite{RBFPNet} methods. 
In lieu of employing a pre-trained optical flow network or an intricate module such as a cost volume, MA-VFI adopts a simple but powerful network for VFI task. Among them, all algorithms are based on deep learning methods. 

\subsection{Results on Single-Frame Interpolation}

To gauge the efficacy of the suggested methodology within the context of single-frame VFI, the evaluation is carried out utilizing the UCF101, Vimeo-90k, and Middlebury datasets. The inputs to MA-VFI consist of a pair of successive frames, denotes as $I_0$, $I_1$, while the resulting output corresponds to the intermediate frame $I_t$. In addition to PSNR, SSIM, and IE metrics, the computation complexity and inference speed are measured in this paper. All techniques are executed on a single GeForce RTX 3090, operating at a resolution of 448×256, to determine their inference speed. Table \ref{table.1} provides a comparison of the evaluation metrics, \emph{e.g.}, PSNR, SSIM, IE, Parameters, and Runtime, between the proposed MA-VFI method and other competitors. The results, obtained from the three public datasets, affirm the efficacy of our approach.

\begin{figure*}[t]
\centering
\includegraphics[width=0.9\textwidth]{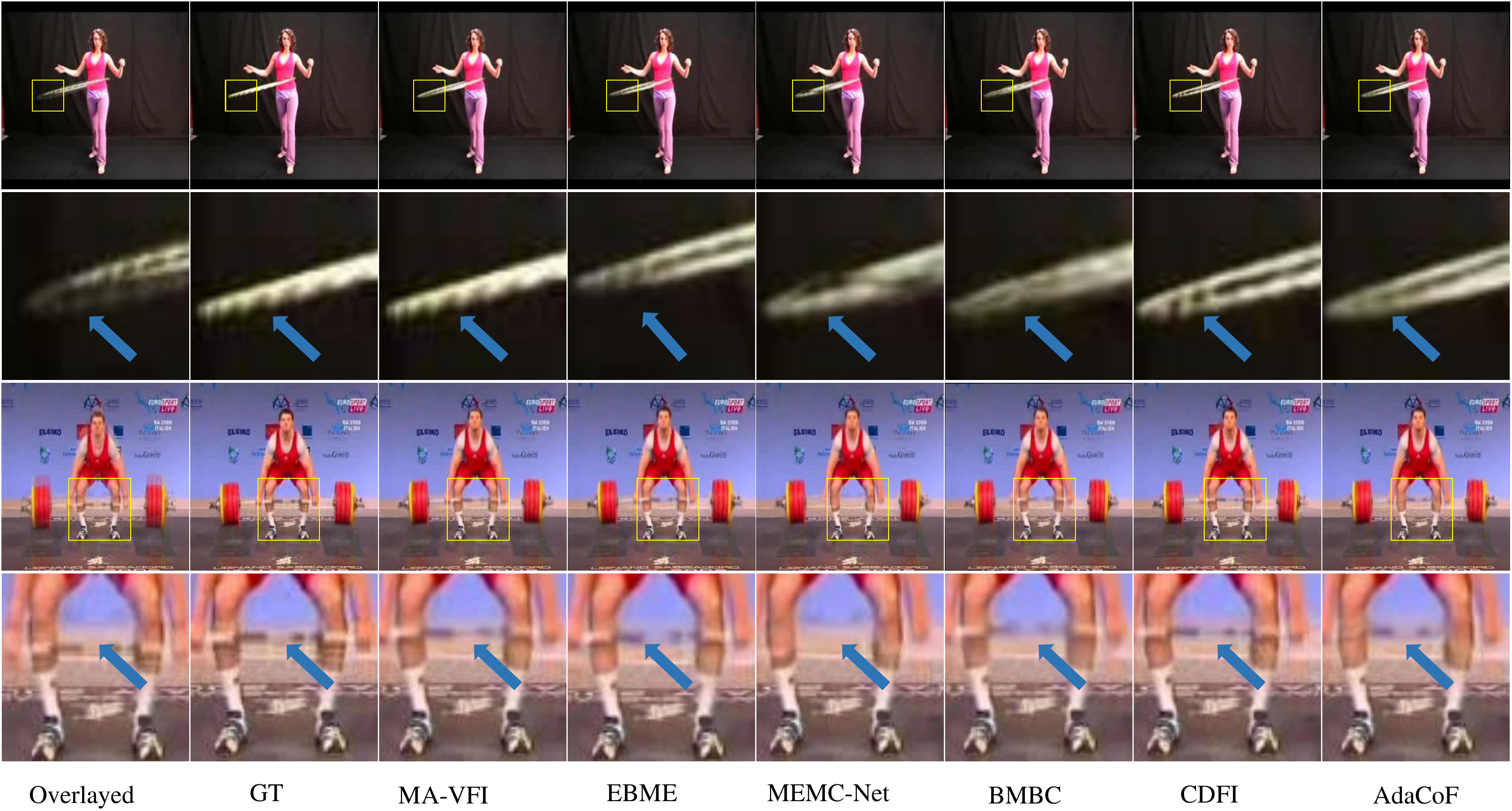} %
\caption{
	Visual juxtaposition of the intermediate frames generated using various approaches within the UCF101 dataset. The superimposed  and the reference frames are positioned by the initial two columns on the left. In the subsequent columns, the uppermost and third rows showcase the intermediate frames produced by distinct methods. The second and fourth rows depict enlarged segments of the yellow rectangular area.
}
\label{fig4}
\end{figure*}

Within the realm of qualitative assessment,
Fig. \ref{fig5} and Fig. \ref{fig4} display the synthesized frames by different methods.
Specifically, the scene of the fast-moving hula hoop and the barbell being lifted instantaneously are visualized in UCF101 dataset. As can be clearly seen in Fig. \ref{fig5}, EBME method,
located in the second row and fourth column, exhibits a phenomenon characterized by the absence of edge details and a blurred appearance.
The proposed MA-VFI, situated in the second row and third column, can effectively preserve the edges of the moving objects and eliminates artifacts in the boundary. 
In Fig. \ref{fig4}, the scene of a small ball with a fast moving and a foot with a large displacement are visualized in Middlebury dataset. The proposed  method MA-VFI achieves better results  on objects with large moving distance. 

Regarding the quantitative assessment, Table \ref{table.1} provides a depiction of the PSNR, SSIM, IE, Parameters, and Runtime assessment metrics in relation to the suggested MA-VFI method and other ones.
The PSNR achieved by the proposed methodology attains a value of 35.31dB on the UCF101 dataset, which is higher 0.24dB than ReMEI-Net~\cite{hucs}. The target of RIFE~\cite{huang2022real} is to achieve a real-time video frame interpolation, with an inference time 5ms faster than the proposed MA-VFI. However, all other metrics are inferior to our method, with a PSNR reduction of 0.35. In addition, the PSNR of ProBoost-Net~\cite{ProBoost-Net} method reaches 36.18dB on the Vimeo-90K dataset. The SSIM reaches 0.979, which is only 0.001 lower than the proposed method MA-VFI. The IE of EA-Net method reaches 2.17 on Middlebury data, which is only 0.26 lower than MA-VFI.  
The distinctive merit of ProBoost-Net lies in its utilization of the ConvLSTM architecture to encapsulate extensive pixel correlations across video frames. 
It is noteworthy to highlight that the suggested approach MA-VFI runs 76\({\times}\) faster than BMBC with a compact and powerful network. While TOFlow \cite{xue2019video} and DVF \cite{liu2017video} algorithms achieve lightweight network, MA-VFI acquires 2.23 dB higher than TOFlow and 1.4dB better than DVF on Vimeo90K dataset in PSNR. While the PSNR value of the SOFT algorithm on the UCF101 dataset is higher than that of the proposed algorithm, the SSIM value is 0.018 lower than the proposed MA-VFI algorithm, and the runtime is 9 times slower than the proposed algorithm. The PSNR value of ProBoost-Net  on Vimeo-90k is 0.22 higher than our proposed algorithm, but the runtime is 15 times slower than our MA-VFI.
The proposed MA-VFI can effectively preserve the edges of the moving objects and eliminates artifacts in the boundary in Fig.~\ref{fig5} and~\ref{fig4}. Overall, Fig.~\ref{fig5}, Fig.~\ref{fig4} and Table \ref{table.1} can collectively demonstrate that by introducing the novel hierarchical  pyramid feature extraction module and cross-scale motion intermediate flow estimation module, low-level and high-level semantic features are extracted from different receptive fields of input frames to capture more motion details of small objects. By directly estimating the intermediate optical flow between adjacent frames, the limitations of intermediate flow calculation are effectively alleviated, achieving a balance between optimal performance and efficiency in the single-frame interpolation task.

\subsection{Results on Multi-frame Interpolation}
	
Within this section, Adobe240-fps dataset is used to evaluate multi-frame interpolation. It consists of 254 sequences, and each one has 9 frames. Specifically, the first and last frames are used to predict three, and five intermediary frames at various time intervals, with $t$ ranging from 0 to 1.

MA-VFI is compared with BMBC \cite{park2020bmbc}, EDSC \cite{cheng2021multiple}, AdaCoF \cite{lee2020adacof}, CAIN \cite{choi2020channel}, and LBEC   \cite{ding2022video} methods.
		 BMBC \cite{park2020bmbc} and EDSC \cite{cheng2021multiple} can produce a frame at a random time point. 
		AdaCoF \cite{lee2020adacof}, CAIN \cite{choi2020channel}, LBEC \cite{ding2022video} limited to interpolating the intermediary frame at timestep t = 0.5.
		Those methods are recursively used to produce  \(\times\)3 and \(\times\)5 results. In particular, we begin by employing the single interpolation method to obtain the intermediate frame \({I_{0.5}}\) from input frames \({I_{0}}\) and \({I_{1}}\), get intermediate frame \({I_{0.25}}\) form \({I_{0}}\) and \({I_{0.5}}\), and then get intermediate frame \({I_{0.75}}\) from \({I_{0.5}}\) and \({I_{1}}\).
		In the same way,  \({I_{0}}\) and \({I_{0.25}}\) are fed into MA-VFI to get intermediate frame \({I_{0.125}}\). It is worth mentioning that we feed the generated intermediate frames \({I_{0.5}}\) and \({I_{0.75}}\) into MA-VFI to get frames \({I_{0.625}}\).

  \begin{table}[ht]
			\renewcommand\arraystretch{1.0}
			\small 
			\caption{Multiple frame video interpolation quantitatively complied with different methodologies across the Adobe 240-fps dataset.
			}
			\centering
			\setlength\tabcolsep{10pt}
			\begin{tabular}{ccccc}
				\hline
				\noalign{\hrule height 1.1pt}
				\multirow{2}{*}{ \textbf{Methods}} & \multicolumn{2}{c}{\textbf{\(\times\)3}}            & \multicolumn{2}{c}{\textbf{\(\times\)5}}           \\ 
				\cmidrule(r){2-3}  	\cmidrule(r){4-5} 
				& \multicolumn{1}{c}{ \textbf{PSNR}}  & { \textbf{SSIM}}  & \multicolumn{1}{c}{ \textbf{PSNR}}  & { \textbf{SSIM}}  \\\hline
				AdaCoF \cite{lee2020adacof}                    & \multicolumn{1}{c}{30.93} &0.946  & \multicolumn{1}{c}{30.58} &0.939  \\ \hline
				CAIN \cite{choi2020channel}                   & \multicolumn{1}{c}{29.52} &\textbf{\textcolor{blue}{0.949}}  & \multicolumn{1}{c}{29.83} &\textbf{\textcolor{blue}{0.950}}  \\ \hline
				BMBC \cite{park2020bmbc}                    & \multicolumn{1}{c}{27.32} &0.921  & \multicolumn{1}{c}{27.44} &0.923  \\ \hline
				EDSC   \cite{cheng2021multiple}                    & \multicolumn{1}{c}{30.82}      &0.947       & \multicolumn{1}{c}{\textbf{\textcolor{blue}{31.15}}}      &\textbf{\textcolor{blue}{0.950}}       \\ \hline
				LBEC   \cite{ding2022video}                    & \multicolumn{1}{c}{\textbf{\textcolor{blue}{31.32}}}      &\textbf{\textcolor{blue}{0.949}}      & \multicolumn{1}{c}{28.16}      &0.887       \\ \hline
				MA-VFI(Ours)                 & \multicolumn{1}{c}{{\textbf{\textcolor{red}{31.56}}}} & {\textbf{\textcolor{red}{0.951}}} & \multicolumn{1}{c}{{\textbf{\textcolor{red}{31.95}}}}      &{\textbf{\textcolor{red}{0.955}}}       \\ \hline
				\noalign{\hrule height 1.1pt}
			\end{tabular}\label{table.2}
		\end{table}

 In this section, Adobe240-fps dataset is used to evaluate multi-frame interpolation. It consists of 254 sequences, and each one has 9 frames. Precisely, the initial and concluding frames are employed to forecast three and five intermediary frames, respectively, spanning diverse time steps denoted as $t$ $\in$ (0,1).

   It is within Table~\ref{table.2}, the comparisons among the values recorded by the various protocols for the PSNR and SSIM values on the Adobe240-fps dataset are furnished.
		The findings underscore the efficacy of the proposed MA-VFI approach in multi-frame interpolation.
		The previous method employs Transformer module to encapsulate extensive pixel correlations across video frames over extended ranges and achieves the better results. However, it requires complicated training steps and complex models. The proposed MA-VFI designs a simple and powerful model to interpolate frames and only costs 0.021s when generating a 448×256 resolution frame. Compared with those, our method achieves comparable visual effects.


To better understand the results on multi-frame interpolation, Fig. \ref{fig6} shows the result of five frames interpolation, which includes EDSC \cite{cheng2021multiple}, LBEC \cite{ding2022video}, and MA-VFI. 
Particularly, the scene is picked on from the frames with large
		motion, \emph{i.e.}, a moving car.
		The results clearly show that
		MA-VFI can better eliminate artifacts and keep the object boundaries in the generated frames.
		In the multi-frame interpolation task, Fig. \ref{fig6} and Table \ref{table.2}
collectively substantiate the superior performance of the proposed MA-VFI method in the context of multi-frame interpolation tasks.

	\begin{figure*}[ht]
		\centering	 \includegraphics[width=1.0\textwidth]{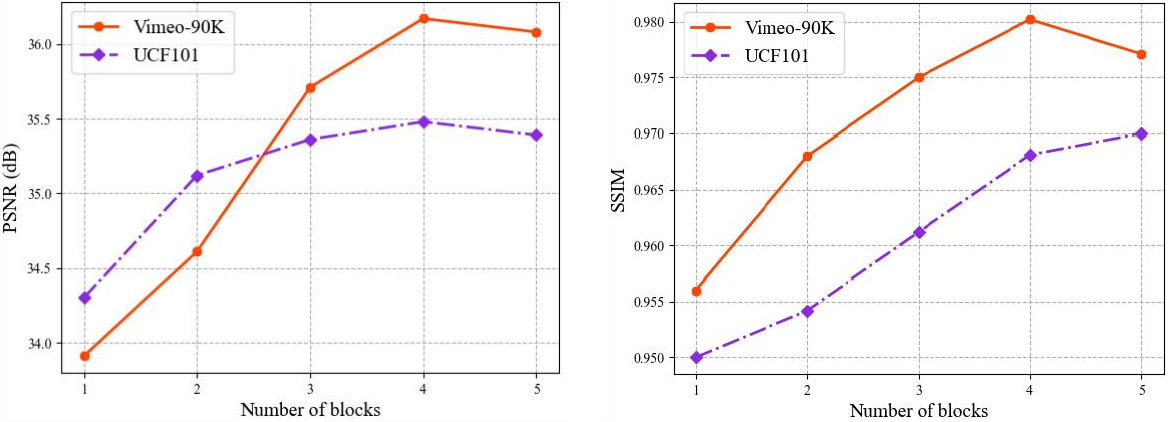}  
				\caption{The PSNR and SSIM metric outcomes from the ablation study, conducted with varying numbers of blocks on the UCF101 and Vimeo-90K datasets, are presented. The left side showcases the PSNR metric results, while the right side illustrates the results for the SSIM metric.}
				\label{fig7}
				\end{figure*}

\subsection{Ablation Studies}
		Within this section, a thorough analysis of the proposed method is conducted through ablation studies. Initially, the examination delves into the impact of the block count on the cross-scale motion structure. Subsequently, a detailed investigation is undertaken to discern the contribution of each individual element within the suggested approach.
		Finally, the influence of the loss function setting on MA-VFI is studied.
	\begin{figure}[ht]
			\centering
			\includegraphics[width=0.5\textwidth]{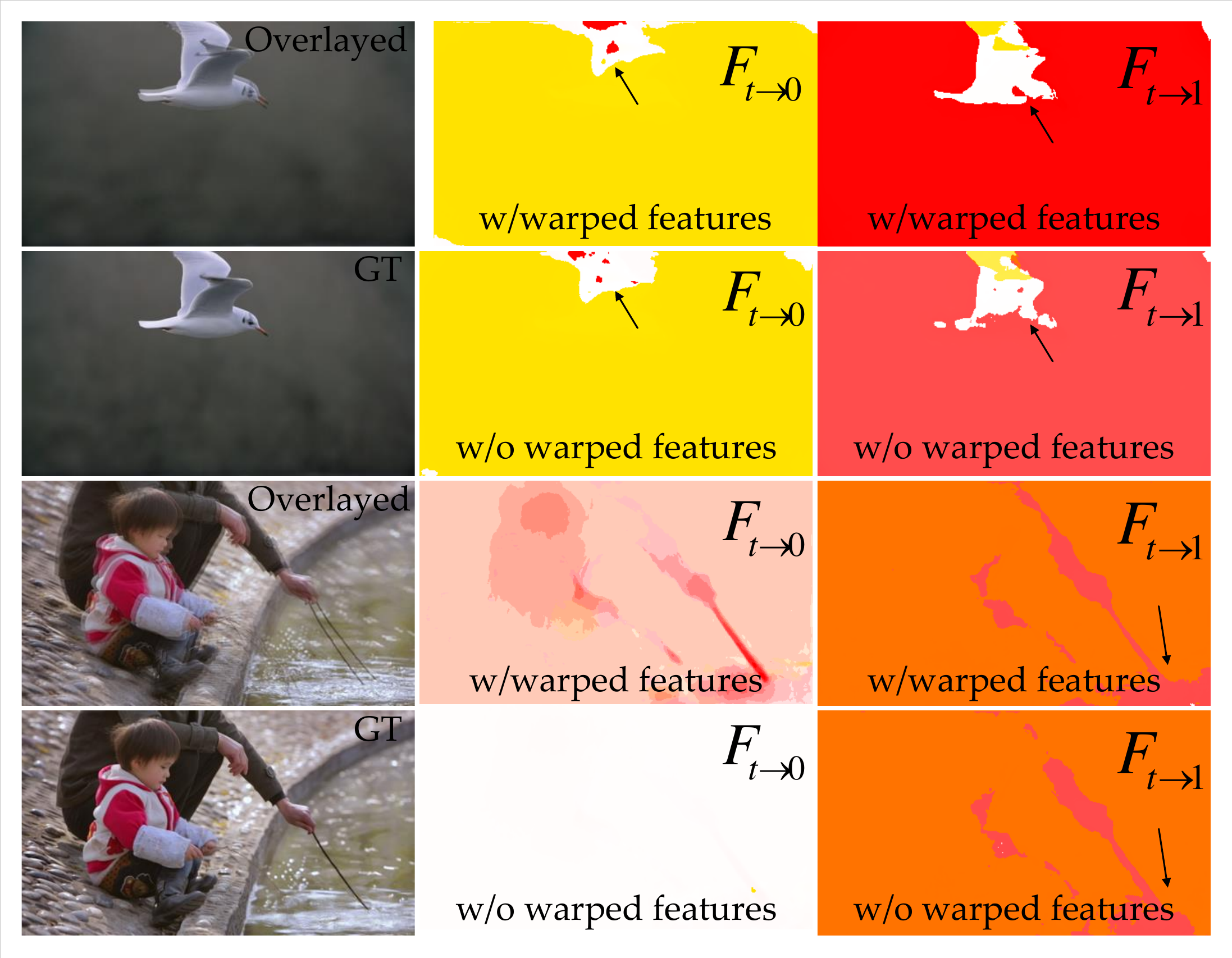}  
			\caption{
				Ablation study on the warped features, the samples presented in the first column consist of overlayed inputs and ground truths. The second and third columns showcase the outcomes of intermediate flow maps, one without warped features and the other with warped features incorporated.
			}
			\label{fig8}
		\end{figure}

  The quality assessment of synthesized intermediate frames across varying numbers of blocks is depicted in Fig.~\ref{fig7}. Specifically, 1, 2, 3, 4 and 5 IFBlocks are used to estimate intermediate flow respectively. 
		They are tested on the UCF101 and Vimeo-90k datasets.
		As observed in Fig.~\ref{fig7}, the outcomes become better as the number of block layers grows from 1 to 4. The result of 4 blocks is better than 5 blocks, thereby thoroughly underscoring the efficacy on the suggested method.

		\begin{table}[ht]
			\renewcommand\arraystretch{0.8}\footnotesize
			\caption{Ablation study outcomes across the UCF101 and Vimeo-90k
				datasets. `FF' means frame features, `IF' means intermediate feature, `FIF' means frame and intermediate features.}
			\centering
			\vspace{0.5em} 
			
			\setlength\tabcolsep{8pt}
			\begin{tabular}{ccccc}
				\hline
				\noalign{\hrule height 1.1pt}
				\multirow{2}{*}{\textbf{Model}}  & \multicolumn{2}{c}{\textbf{UCF101}}        & \multicolumn{2}{c}{\textbf{Vimeo-90K}}    \\ 
				\cmidrule(r){2-3}  	\cmidrule(r){4-5}
				& \multicolumn{1}{c}{\textbf{PSNR}}  & {\textbf{SSIM}}  & \multicolumn{1}{c}{\textbf{PSNR}}  & {\textbf{SSIM}} \\ \hline
				MA-VFI w/o FF   & \multicolumn{1}{c}{35.2122}&0.9691 & \multicolumn{1}{c}{35.5314} & 0.9780      \\ \hline
				MA-VFI w/o IF  & \multicolumn{1}{c}{35.2964}&0.9693       & \multicolumn{1}{c}{35.9436}&0.9797      \\ \hline
				MA-VFI w/o FIF  & \multicolumn{1}{c}{35.1136}&0.9685      & \multicolumn{1}{c}{34.9803}&0.9751      \\ \hline
				MA-VFI w/o residual           & \multicolumn{1}{c}{35.2997} & 0.9693     & \multicolumn{1}{c}{35.9227}&  0.9797    \\ \hline
				MA-VFI w/o $\mathcal{L}_{flow}$       & \multicolumn{1}{c}{35.3014}&0.9693     & \multicolumn{1}{c}{35.9291}&0.9797       \\ \hline\hline
				MA-VFI(Ours)                 & \multicolumn{1}{c}{35.3075} & 0.9693 & \multicolumn{1}{c}{35.9583} & 0.9798 \\ \hline
				\noalign{\hrule height 1.1pt}
			\end{tabular}\label{table.3}
		\end{table}

		Within MA-VFI, an effect on the interactions between the extracted features and intermediate flow maps in different layers is verified.
		Recall that the higher features are used to estimate the intermediate flow, then warps the lower features by predicted flow maps for spatial alignment. The warped features and current flow maps are used to estimate the next level intermediate flow. The warped features include two parts: frame features $F^i_{t0}$, $F^i_{t1}$ and intermediate feature $F^i_{t}$ in Eq. \ref{frame features} and Eq. \ref{Intermediate features}. We build a model by removing the frame and intermediate features respectively, while the rest of the model remains unchanged. In Table \ref{table.3}, from first to third rows, it shows that the warped features promote the mutual assistance between input frames features and flow maps during interpolating. The frame features assume a pivotal role by aligning the flow maps with the extracted features, thereby contributing significantly to the overall process.

		To further demonstrate the importance of the warped features, Fig. \ref{fig8} offers visualizations of intermediate flow maps, comparing the outcomes with and without warped features. 
		As can be seen from Fig. \ref{fig8}, the model incorporating warped features yields more intricate and detailed intermediate flow maps.
		Furthermore, the residual concatenation is employed to optimize flow and guide map. We also build a new model by removing the residual concatenation. From the fourth row in Table \ref{table.3},  residual compensation is advanced for VFI performance.
		Overall, Table \ref{table.3} and Fig. \ref{fig8} jointly verify the effectiveness of  cross-scale motion intermediate flow estimation model.
		By combining the extracted features with the intermediate flow map, nonlinear motion is better represented.
		We have already mentioned the addition of an intermediate flow-directed loss to MA-VFI. This loss is used to provide more supervision for the frame synthesis process.

		As part of our effort to validate the impact of the intermediate flow-directed loss over the model, we have compared the values assigned to indicators of PSNR and SSIM evaluation before and after the loss has been removed from the model.
Therefore, as illustrated in Table \ref{table.3}, removing the intermediate flow-directed loss and smooth loss undermines the performance significantly.
	Overall, the Table \ref{table.3}, Fig. \ref{fig8}, and Fig.~\ref{fig7} can collectively showcase  the pre-eminence and efficacy afforded by the MA-VFI method.

\section{Limitations and Future Works}

In existing research, although our algorithm has achieved a certain balance between model inference time and frame interpolation performance, there is still room for improvement compared to the RIFE algorithm. In industrial applications, video interpolation models are typically deployed on hardware devices to increase video frame rates and enhance video quality. However, current high-performance video interpolation algorithms often rely on GPU computing power and cannot be deployed on hardware devices with lower computing power. Therefore, designing video interpolation algorithms with low computational complexity and lightweight characteristics is an urgent problem in the industry to meet practical application needs. Additionally, on the Vimeo-90k dataset, ProBoost-Net achieved the best PSNR value, far superior to the proposed MA-VFI. Therefore, future research will focus on two aspects: first, efforts to reduce the inherent complexity of the model to improve algorithm efficiency and performance; second, ensuring that while simplifying the model, the sacrifice to its effectiveness is minimized. Furthermore, we are considering conceptualizing a separate network specifically designed to support a range of different scales of image optical flow estimation and extending this approach to the field of video interpolation.

 \section{Conclusions} \label{Conclusion}
		In this paper, we put forward MA-VFI network designed for the task of video frame interpolation (VFI). It directly estimates intermediate flow maps, which can effectively captures the nuances of nonlinear motion observed in real-world scenarios. Specifically, a pyramid feature module and a cross-scale motion structure make up MA-VFI. In the pyramid feature module, the low and high features are extracted from given frames on different receptive fields. This makes it easier to capture complex motion. The
		extracted features are used to estimate and refine intermediate
		flow maps by a cross-scale motion structure. Specifically, it
		warps the lower features by higher flow maps for spatial
		alignment, and then employs the warped features to estimate
		the next level intermediate flow. This can promote the mutual
		assistance between input frames features and flow maps during
		interpolating frames. 
		Lastly, an intermediate flow directional loss has been artfully designed to provide precise guidance for the estimation of intermediate flow. This aids the model in mitigating challenges such as image blurring and the occurrence of spurious artifacts. Practically, the experiments have substantiated the  pre-eminence and efficacy of suggested method.

\bibliographystyle{elsarticle-num}  
\bibliography{mybibfile}

\end{document}